\documentclass{article}
\usepackage{spconf}
\usepackage{%
algorithm,%
algpseudocode,%
amsfonts,%
amsmath,%
amssymb,%
amsthm,%
bbm,%
caption,%
cite,%
comment,%
extarrows,%
float,%
graphicx,%
hyperref,%
paralist,%
pgfplots,%
wrapfig,%
subcaption,%
tikz,%
url,%
xcolor,%
}
\usepackage{listings}
\usepackage{indentfirst}

\usepackage{epstopdf}

\usepackage{todonotes}

\pgfplotsset{plot coordinates/math parser=false}
\usepgflibrary{%
patterns,%
}

\usetikzlibrary{%
arrows,%
automata,%
chains,%
decorations.pathreplacing,%
intersections,%
positioning,%
shapes,%
}

\theoremstyle{definition}

\newtheorem{theorem}{Theorem}

\theoremstyle{remark}

\definecolor{darkgreen}{rgb}{0.0, 0.5, 0.0}

\newcommand{\set}[1]{\left\lbrace#1\right\rbrace}

\newcommand{\E}{\mathbb{E}}

\newcommand{\R}{\mathbb{R}}

\newcommand{\cA}{\mathcal{A}}

\newcommand{\cE}{\mathcal{E}}

\newcommand{\cG}{\mathcal{G}}

\newcommand{\cN}{\mathcal{N}}

\newcommand{\cV}{\mathcal{V}}

\renewcommand{\ge}{\geqslant}
\renewcommand{\le}{\leqslant}

\renewcommand{\ge}{\geqslant}
\renewcommand{\le}{\leqslant}

\title{Decentralized Learning Strategies for Estimation Error Minimization with Graph Neural Networks}

\name{Xingran Chen$^{\star \dagger}$ \qquad Navid NaderiAlizadeh$^{\star}$ \qquad Alejandro Ribeiro$^{\dagger}$ \qquad Shirin Saeedi Bidokhti$^{\dagger}$
\thanks{This work was supported by National Science Foundation of China (Grant No. 62401111), NSF AI Institute for Learning-Enabled Optimization at Scale (Grant No.~NSF-CCF-2112665), and NSF CAREER Award (Grant No.~NSF-CCF-2047482).}
\thanks{This work has been accepted to IEEE ICASSP 2026.}
}
\address{$^{\star \dagger}$ Rutgers University,
$^{\star}$ Duke University,
$^{\dagger}$ University of Pennsylvania}

\begin{document}

\maketitle
\begin{abstract}
We address real-time sampling and estimation of autoregressive Markovian sources in dynamic yet structurally similar multi-hop wireless networks. Each node caches samples from others and communicates over wireless collision channels, aiming to minimize time-average estimation error via decentralized policies. Due to the high dimensionality of action spaces and complexity of network topologies, deriving optimal policies analytically is intractable. To address this, we propose a graphical multi-agent reinforcement learning framework for policy optimization. Theoretically, we demonstrate that our proposed policies are transferable, allowing a policy trained on one graph to be effectively applied to structurally similar graphs. Numerical experiments demonstrate that (i) our proposed policy outperforms state-of-the-art baselines; (ii) the trained policies are transferable to larger networks, with performance gains increasing with the number of agents;  (iii) the graphical training procedure withstands non-stationarity, even when using independent learning techniques; and (iv)  recurrence is pivotal in both independent learning and centralized training and decentralized execution, and improves the resilience to non-stationarity.
\end{abstract}
\begin{keywords}
Decentralized decision making, multi-agent reinforcement learning, graph neural networks, transferability, estimation
\end{keywords}
\section{Introduction}\label{sec: Introduction}

Accurate and timely estimation of physical processes is essential for applications such as IoT sensing, robot swarm coordination, and autonomous driving \cite{correlatedAoI, ETLBDM2021}. A fundamental difficulty is that minimizing real-time estimation error critically depends on timely communication. In this paper, we study real-time sampling and estimation in \textit{dynamic multi-hop} wireless networks. Each node (e.g., device) observes a physical process modeled as a Gauss–Markov source \cite{cxr2020estimation, correlatedAoI}, and must maintain accurate, fresh estimates of all other nodes' states. Building on the fundamental difficulty above, we identify three major challenges: (i) Joint timeliness and estimation: real-time information must be collected and used immediately for estimation, rather than being fully processed and stored before reliable transmission. This coupling between freshness and accuracy requires new co-design strategies. (ii) Dynamic network topology: Node mobility, link variations, and failures continuously alter network connectivity. (iii) Decentralized decision-making: Centralized operation is infeasible in large-scale networks. Nodes must act based on local observations.

To tackle the first challenge, the concept of Age of Information (AoI) was introduced in 2011 \cite{5984917} as a metric to capture timeliness. AoI has been widely used as a proxy for real-time estimation error \cite{YSunestimation2020, YSYPEUB2020TIT, cxr2020estimation}. Smaller AoI typically yields lower estimation error. This link has motivated studies on estimation through the lens of AoI \cite{YSYPEUB2017ISIT, YSYPEUB2020TIT, YSunestimation2020, 9589639, SSHSMVBSCRM2022CL, cxr2020estimation}, though most focus on one-hop settings, leaving applicability to complex topologies unclear. To tackle the second challenge, optimal transmission policies for freshness and estimation have been investigated in multi-hop networks, but mainly under centralized scheduling and fixed graphs \cite{SFAGKDRB2019, BBASSU2019}. Even when decentralized strategies are considered, such as the randomized policy in \cite{VTRTEM2022ToN}, they remain simplistic, while more advanced designs still rely on central control \cite{VTRTEM2022ToN, NicholasJones2023}. Thus, near-optimal decentralized mechanisms for dynamic multi-hop networks remain an open problem. To tackle the third challenge, learning-based approaches are utilized. Multi-agent reinforcement learning (MARL) has been effective in coordinating agents \cite{epymarl}, but classical MARL methods degrade under evolving topologies. This motivates incorporating graph neural networks (GNNs) \cite{FGamaStability2020, LuanaRuiz2020}, whose permutation equivariance and transferability enable a graphical MARL framework suitable for dynamic networks.

In this work, we study decentralized sampling and estimation of $M$ Gauss–Markov sources over wireless collision channels in \textit{dynamic yet structurally similar} multi-hop networks. Each node decides when to sample, whom to transmit to, and what to transmit, aiming to minimize time-average estimation error. We first show that, when decisions are process-independent, minimizing estimation error is equivalent to minimizing AoI. The main contributions are:
\begin{compactenum}[1)] 
\item We propose a novel designed graphical MARL framework that integrates a graphical actor, a graphical critic, and an action distribution operator to jointly determine when to sample, whom to transmit to, and what to transmit. A key feature of the framework is its \textit{transferability}. 
\item  We develop a theoretical framework that establishes transferability of graphical MARL policies across dynamic but structurally similar networks. Unlike existing GNN transferability studies, our results apply to a MARL framework where GNNs are only a component, and therefore cannot be derived from GNN-only analyses.
\item Extensive experiments demonstrate that (i) the proposed graphical MARL  outperforms classical MARL, while the centralized training with decentralized execution (CTDE) mitigates non-stationarity relative to independent learning; (ii) graphical MARL exhibits strong transferability, yielding performance gains in large-scale networks; and (iii) recurrence enhances resilience to non-stationarity in both CTDE and independent learning.
\end{compactenum}

\section{System Model}\label{sec:systemModel}

Consider $M$ statistically identical nodes communicating over a {\it connected} undirected graph. Let $\cG = (\cV, \cE)$ denote the graph, where $\cV = \set{1,2,\cdots,M}$ is the node set and $\cE$ is the edge set representing communication links. For each node $i\in\cV$, define its neighborhood as $\partial_i = \set{j \mid (i,j)\in\cE}$. The graph $\cG$ may evolve over time\footnote{Since the graph $\cG$ may vary with time $t$, it should more precisely be denoted by $\cG_t$. For notational simplicity, we will instead use $\cG$.}; however, we assume that successive graphs remain {\it structurally similar}\footnote{Here, {\it similarity} means that the graphs are sampled from the same graphon \cite[Definition~3]{GNNerr}. This assumption is broad but practical: nodes may move or the network may scale while preserving average degree, as often seen in wireless networks (mobile ad hoc, vehicular, UAV/swarm), where instantaneous topology changes but large-scale statistical structure stays stable.}.

Let time be slotted. Each node $i$ observes a physical process $\{\zeta_{k}^{(i)}\}_{k \ge 0}$ in slot $k$, $\zeta_{k+1}^{(i)} = \zeta_{k}^{(i)} + \Lambda_{k}^{(i)}$,
where $\Lambda_{k}^{(i)}\sim\cN(0,\sigma^2)$ are i.i.d. across all $i$ and $k$ \cite{cxr2020estimation, correlatedAoI}. The physical processes are mutually independent across nodes, and we set $\zeta_{0}^{(i)}=0$ for all $i\in\mathcal{V}$.  
Each node is equipped with a cache: it can store samples of its own process and packets received from neighbors. In the beginning of each slot, a node decides whether to transmit a packet or remain silent. If transmitting, the node selects one packet from its cache and one neighbor $j\in\partial_i$ as the receiver. 
The communication medium is modeled as a collision channel: If two or more nodes simultaneously transmit to a common neighbor, or if two nodes on opposite ends of an edge simultaneously transmit to each other, all involved transmissions fail. Only the sender observes the collision feedback. Packet delivery is assumed to take one slot. 

Based on the received packets, each node can estimate the processes of others. Let $\hat{\zeta}_{j, k}^{(i)}$ denote the estimate of $\zeta_{k}^{(j)}$ in time slot $k$ from the perspective of node $i$. By convention, let $\hat{\zeta}_{i, k}^{(i)}=\zeta_{k}^{(i)}$ for all $i$ and $k$, and $\hat{\zeta}_{j, 0}^{(i)}=0$ for all $i, j\in\mathcal{V}$. We utilize the MMSE (Kalman) estimator due to its optimality \cite{correlatedAoI}: $\hat{\zeta}_{j,k}^{(i)} = \mathbb{E}[\zeta_{k}^{(j)}\mid \zeta_{\tau_{i,j}}^{(j)}]$,
where $\zeta_{\tau_{i,j}}^{(j)}$ denotes the latest packet from node $j$ that is cached at node $i$. We define the average sum of estimation errors (ASEE) as our performance metric: 
{\small
\begin{align*}
L^{\pi}=\lim_{K\rightarrow\infty}\E[L_K^{\pi}],\,\, L_K^{\pi}=\frac{1}{M^2K}\sum_{k=1}^{K}\sum_{i, j=1}^{M}\big(\hat{\zeta}_{j,k}^{(i)} - \zeta_{k}^{(j)}\big)^2
\end{align*}
}
where  $\pi\in\Pi$ refers to a decentralized sampling and transmission policy, and $\Pi$ is the set of all possible policies. Thus, we need to solve the optimization problem: $\min_{\pi\in\Pi}\,\, L^{\pi}$.

A sampling and transmission policy is defined as 
$\pi=\{\mu_{k}^{(i)}, \nu_{k}^{(i)}\}_{i\in\cV, k\ge0}$. At time $k$, if $\mu_{k}^{(i)}\in\partial_i$, then node $i$ transmits a packet in its cache that originated from node~$\mu_{k}^{(i)}$, and sends it to neighbor $\nu_{k}^{(i)}\in\partial_i$; if $\mu_{k}^{(i)} = i$, then node~$i$ remains silent in slot $k$; and we conventionally set $\nu_{k}^{(i)}=i$.  

We consider two general classes of policies: \textit{oblivious} policies and \textit{non-oblivious} policies. In the former class, decision-making is independent of the processes being monitored. In the latter class, decision-making depends on the processes. From \cite[Lemma~1]{GNNerr}, we prove that, in the class of \textit{oblivious} policies, minimizing the ASEE is equivalent to minimizing the time-average AoI.

\section{Proposed Graphical MARL Framework}\label{sec: GNNRL}

\begin{figure}
\centering \includegraphics[width=0.475\textwidth]{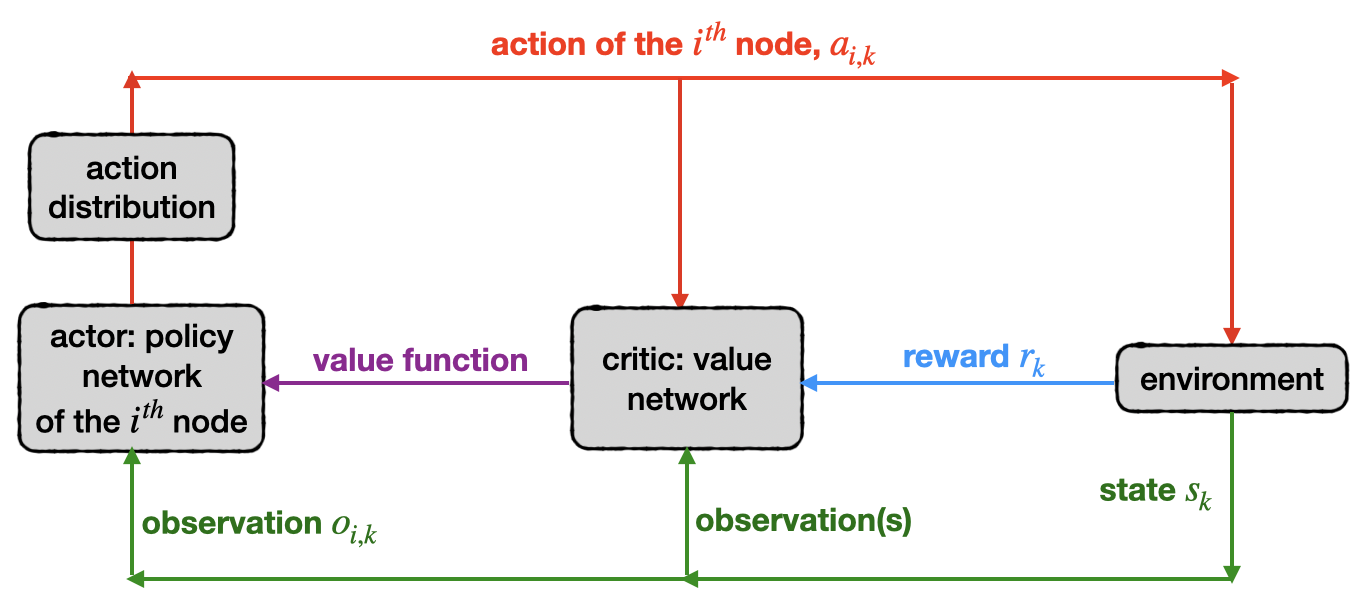} 
\caption{The proposed graphical MARL framework.}
\label{fig_outline} 
\end{figure}

In this section, we present the proposed MARL framework. We begin with a general overview, as shown in Fig.~\ref{fig_outline}, followed by discussions of two key components: (i) graphical actors and critics (see Section~\ref{sec: Graph Actor and Critic}), and (ii) the action distribution (see Section~\ref{sec: Action Distribution}). In Section~\ref{subsec: Permutation Invariance and Transferability}, we provide theoretical results.

After obtaining its observation $o_{i, k}$, node $i$ makes a decision $(\mu_{k}^{(i)}, \nu_k^{(i)})$ based on a combination of a policy network (actor) and an action distribution. 
Let the policy network of node $i$ be denoted by $\pi(o_{i,k};\theta_i)$, and let the associated action distribution be denoted by $\cA(\cdot;\vartheta_i)$. 
Since all nodes are homogeneous, we employ \emph{parameter sharing} as in prior MARL work~\cite{TabishRashid2018}, such that $\theta_1=\cdots=\theta_M=\theta$ and $\vartheta_1=\cdots=\vartheta_M=\vartheta$. 
Thus, the decisions of node $i$ at time $k$ is $(\mu_{k}^{(i)}, \nu_{k}^{(i)}) \sim \cA\!\left(\pi(o_{i,k}; \theta);\vartheta\right)$.

To evaluate the effectiveness of actions, we introduce a value network (critic). 
The critic estimates the expected return given the current state, and its feedback is used to update the policy network (actor). 
We adopt two well-known actor-critic frameworks: IPPO and MAPPO~\cite{epymarl, Volodymyr2016}.%
\footnote{Our framework builds on Proximal Policy Optimization (PPO). Recent studies show that MAPPO, a PPO-based variant, is competitive with or outperforms off-policy methods such as MADDPG, QMix, and RODE in both sample efficiency and wall-clock time \cite{computationexplexity}.}
In IPPO, each node independently trains its own actor–critic pair from local observations.  In MAPPO, nodes share a centralized critic that leverages the joint observations, while keeping the actors decentralized.

At the end of time $k$, the environment provides a reward $r_k$, defined as $r_k = - \frac{1}{M^2}\sum_{i, j\in\cV}\big(\zeta_{k}^{(j)}-\hat{\zeta}_{j,k}^{(i)}\big)^2$.
This reward corresponds to the average estimation error across all nodes, motivated by two factors: (i) all nodes are homogeneous, and (ii) they cooperate to minimize the ASEE.

\subsection{Graphical Actor and Critic}\label{sec: Graph Actor and Critic}

Since nodes in a wireless network naturally form a graph topology, a graph-based structure for the learning model is appropriate. 
For the actor, we adopt graph recurrent neural networks (GRNNs), which are provably permutation-equivariant \cite[Proposition~1]{Transferability}, and have been shown in simulations to outperform both GNNs and RNNs.
For the critic, we use standard GNNs without recurrence, yielding higher computational efficiency than GRNNs. Details of the actor and critic graph representations and signals are given in \cite[Section~IV-B]{GNNerr}.

\subsection{Action Distribution}\label{sec: Action Distribution}
To select an action $(\mu_{k}^{(i)}, \nu_{k}^{(i)})$, each actor outputs $\hat{y}^{(i)} = \pi(o_{i,k};\theta)\in\R^{M\times G}$, which is then passed to the action distribution $\cA(\cdot; \vartheta)$, Specifically, 
\begin{align}\label{eq:ActionDist1}
(\mu_{k}^{(i)}, \nu_{k}^{(i)}) \sim \cA(\hat{y}^{(i)}; \vartheta)\triangleq F_\text{softmax}\left(\hat{y}^{(i)} \vartheta (\hat{y}^{(i)})^T\right), 
\end{align}
where $\vartheta\in\mathbb{R}^{G\times G}$ is a learnable parameter matrix. 
Importantly, the number of parameters in $\vartheta$ depends only on the output feature dimension $G$, and is independent of the number of nodes $M$.

\subsection{Fundamental Analysis: Transferability}\label{subsec: Permutation Invariance and Transferability}
Given a graphon $W$ and a sequence of node labels $\{u_i\}_{i=1}^m$, let $\Xi$ denote the finite graph sampled from $W$, and let $\Xi_m$ denote the associated induced graphon via $\Xi$ \cite[Section~III-C]{GNNerr}.
Graphon filters $T_{B,W}$, $T_{C,W}$, and $T_{D,W}$ are defined in \cite[Eqn.~(23)]{GNNerr}. Based on these operators, the graphon recurrent neural network (WRNN), denoted by $\Psi(\cdot)$ \cite[Eqn.~(27)]{GNNerr}, is constructed. The WRNN can be interpreted as the continuous analogue of the actor $\pi(\cdot;\theta)$.
Throughout this subsection, we assume that the convolutional filters in the WRNN satisfy \cite[Assumption~1]{GNNerr}, and that the activation functions $\rho_1$ and $\rho_2$ in the WRNN satisfy \cite[Assumption~2]{GNNerr}.

\textit{1) Transferability in GRNNs}:
Since GNNs possess transferability \cite{Transferability}, it is natural to expect GRNNs, as their extension, to inherit this property. 
Let $\{X_t\}_{t=1}^T$ denote graphon signals, and let $\{x_{t,m}\}_{t=1}^{T}$ be the corresponding induced graph signals; conversely, the finite signals $\{x_{t,m}\}_{t=1}^{T}$ can be embedded back into the graphon signals $\{X_{t,m}\}_{t=1}^{T}$ \cite[Section~III-C]{GNNerr}. 
We denote 
\begin{align}
Y&=\Psi(T_{B}, T_{C}, T_{D}; W, \{X_t\}_{t=1}^T),\label{eq:PsiY}\\
Y_m&=\Psi(T_{B}, T_{C}, T_{D}; W_{\Xi_m}, \{X_{t,m}\}_{t=1}^T).\label{eq:PsiYm}
\end{align}

\begin{theorem}[transferability in GRNNs]\label{thm: WRNN approximation on a generic graph}
Let $Y$ and $Y_m$ be defined in \eqref{eq:PsiY} and \eqref{eq:PsiYm}.  Assume the input and output feature dimensions satisfy $F=G=1$, and define
$\eta_1=\max_{1\le t\le T}\|X_t\|$ and $\eta_2=\max_{1\le t\le T}\|X_t-X_{t,m}\|$.
Then, for any $0<\epsilon\leq 1$, it holds that\footnote{As noted by \cite{SMRLGK2023}, this norm is $\|\cdot\|_{L^2[0, 1]}$.}
\begin{align}\label{eq: extend WNN approximation on a generic graph2}
\|Y - Y_m\|\leq\frac{T(1+T)}{2} (\Theta_1+\Theta_3)\eta_1 +T \Theta_2\eta_2,
\end{align}
where $\Theta_1 = (\Omega+\frac{\pi\kappa^\epsilon_{W_{\Xi_m}}}{\delta^\epsilon_{WW_{\Xi_m}}})\|W - W_{\Xi_m}\|$,  $\Theta_2 = \Omega\epsilon + 2$, and $\Theta_3 = 2\omega\epsilon$.
\end{theorem}

Theorem~\ref{thm: WRNN approximation on a generic graph} demonstrates that the output of a WRNN can be approximated by a GRNN on a large graph sampled from the same graphon. The approximation error consists of three main components: (i) $\frac{T(1+T)}{2}\Theta_1\eta_1$ (graph sampling error), which decreases as the distance between the sampled graph and the graphon ($\Theta_1$) becomes smaller; (ii) $T\Theta_2\eta_2$ (signal sampling error:), which decreases when the sampled graph signal better approximates the graphon signal; (iii) $\frac{T(1+T)}{2} \Theta_3\eta_1$ (filter design error), which can be reduced by choosing convolutional filters with smaller parameters $\epsilon$ or $w$, as described in \cite{Transferability}. Transferability property holds only for convolutional filters built on graph filters \cite{Transferability}, such as GCNConv, TAGConv \cite{GNNsheet}.

$\|Y - Y_m\|$ depends on the recurrence depth $T$, since errors introduced at each step propagate forward and accumulate across subsequent steps. Summing these gives $\sum_{t=1}^T t=\frac{T(1+T)}{2}$, which explains the $\frac{T(1+T)}{2}$ term in the bound.

\textit{2) Transferability in the Action Distribution:}
We now turn to the transferability of the action distribution. Since action distributions are discrete, we use a graphon-based approach and define their \textit{limit action distribution} \cite[Section~V-B]{GNNerr} as the key tool in the proof. The matrix $\vartheta\in\R^{G\times G}$ in \eqref{eq:ActionDist1} introduces a graphon operator $T_{W_{\vartheta}}$ \cite[Section~V-B]{GNNerr}.  Given graphon signals $\{X^{(1)}_t\}_{t=1}^{T}$, $\{X^{(2)}_t\}_{t=1}^{T}$ and their induced counterparts $\{X_{t,m}^{(1)}\}_{t=1}^{T}$, $\{X_{t,m}^{(2)}\}_{t=1}^{T}$, we construct the limit action distributions $\tilde{\cA}$ and $\tilde{\cA}_m$ as continuous relaxations of the discrete action distribution $\cA$ \cite[Section~V-B]{GNNerr}. For $j\in\{1,2\}$, by replacing $\{X_t\}_{t=1}^T$ in \eqref{eq:PsiY} with $\{X_t^{(j)}\}_{t=1}^T$, and $\{X_{t,m}\}_{t=1}^T$ in \eqref{eq:PsiYm} with $\{X_{t,m}^{(j)}\}_{t=1}^T$, we obtain $Y^{(j)}$ and $Y_m^{(j)}$, respectively.

\begin{theorem}[Transferability in action distributions]\label{thm: action distribution A}  
For $j\in\set{1,2}$, let $Y^{(j)}$ and $Y^{(j)}_m$ be defined above. Define $\eta_3=\max_{j\in\set{1,2}}\|Y^{(j)}-Y_m^{(j)}\|$. Then for any $0<\epsilon\le 1$,  
\begin{align}\label{eq: action distribution-A}   
&\left|\tilde{\cA}(Y^{(1)}, Y^{(2)}) - \tilde{\cA}_m(Y^{(1)}_m, Y^{(2)}_m)\right| \nonumber \\
&\le \Gamma \|T_{W_\vartheta}\|\left(\|Y^{(2)}\|
+ \|Y_m^{(1)}\|\right)\eta_3,   
\end{align}  
where $\Gamma$ is a constant independent of the WRNN.  
\end{theorem}  	

From Theorem~\ref{thm: WRNN approximation on a generic graph}, the output discrepancy $\eta_3$ can be made small by properly designing the WRNN. Hence, Theorem~\ref{thm: action distribution A} implies that, for any given pair of input sequences $\{X_t\}_{t=1}^{T}$ and $\{X_{t,m}\}_{t=1}^{T}$, the \textit{pointwise} distance between the limit action distributions, i.e., $\cA(Y^{(1)}, Y^{(2)})$ and $\cA_n(Y^{(1)}_m, Y^{(2)}_m)$, can be small. Combining Theorems~\ref{thm: WRNN approximation on a generic graph} and~\ref{thm: action distribution A}, we conclude that the proposed framework is transferable: models trained on smaller graphs generalize effectively to larger ones. Moreover, the error bound improves with network size (or density). This requirement is naturally satisfied in IoT and 6G wireless networks where the abundance of devices ensures reliable generalization.

\section{Experimental Results}\label{sec: Simulations}

\subsection{Experimental Setup and Baselines}\label{subsec: Experimental Setup}
	
We evaluate the proposed algorithms on both synthetic and real networks. (i) Synthetic networks: we consider Watts–Strogatz graphs and stochastic block models\footnote{The models are widely used in practice: the Watts–Strogatz graph model is applied in heterogeneous sensor networks \cite{WSpractice}, while the stochastic block model is used for community detection in large-scale networks \cite{SBMpractice}.}, each with $N=10$ nodes.\footnote{Experiments fail if $M\geq12$ due to limited computational (GPU) resources.}. (ii) Real-world network. We use the {\it aus\_simple} topology from \cite{topologyzoo}, consisting of $7$ connected nodes. The topology remains fixed, but node indices are randomly permuted at the start of each learning episode, which corresponds to a special case of similar graphs with unchanged structure but relabeled nodes.

For both synthetic and real-world networks, each learning episode lasts $1024$ steps, with $3000$ episodes in total. In the synthetic case, a new graph of the same type is generated at the start of each episode; in the real-world case, node re-numbering introduces variability despite the fixed topology. 
To evaluate the learning-based models during training, we interrupt the process every $10$ episodes to assess the model on a set of $30$ test episodes or graphs.

We compare against three baselines: (i) {\it Classical MARL} -- classical IPPO and MAPPO implementations from \cite{epymarl}; (ii) {\it Uniform transmitting} -- each node with cached packets transmits to adjacent nodes with equal probability; and (iii) {\it Adaptive age-based policy} -- each node with cached packets transmits those with smaller age with higher probability. Further details are provided in \cite[Section~VI-B]{GNNerr}.

\subsection{Numerical Results}\label{sec: Numerical Results}	
\begin{figure}[htbp]
\centering
\begin{subfigure}[b]{0.495\linewidth}
    \centering
    \includegraphics[height=3.3cm]{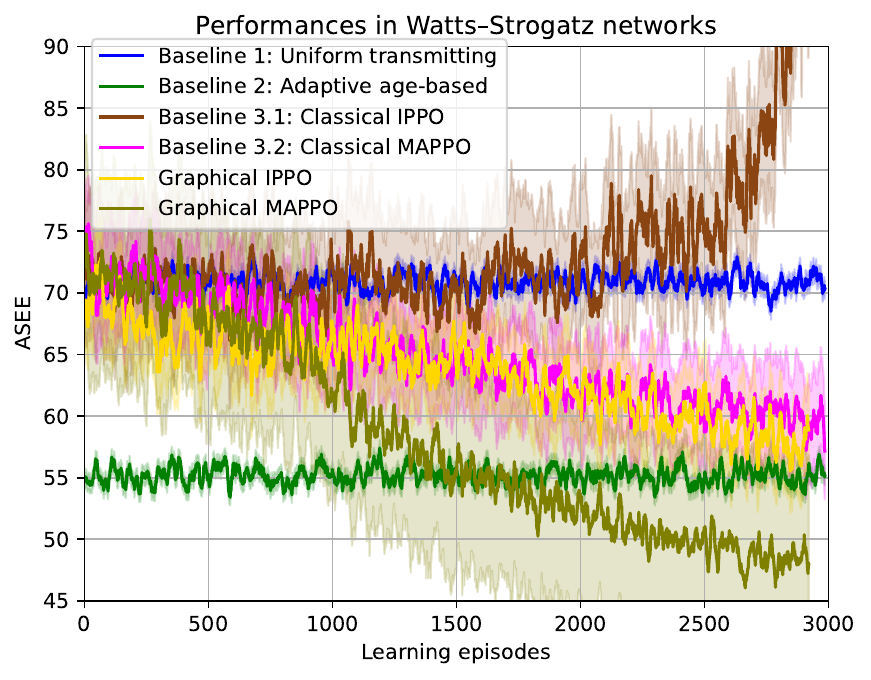}
    \caption{Watts–Strogatz networks}
\end{subfigure}
\begin{subfigure}[b]{0.495\linewidth}
    \centering
    \includegraphics[height=3.3cm,trim=0 0 0 18,clip]{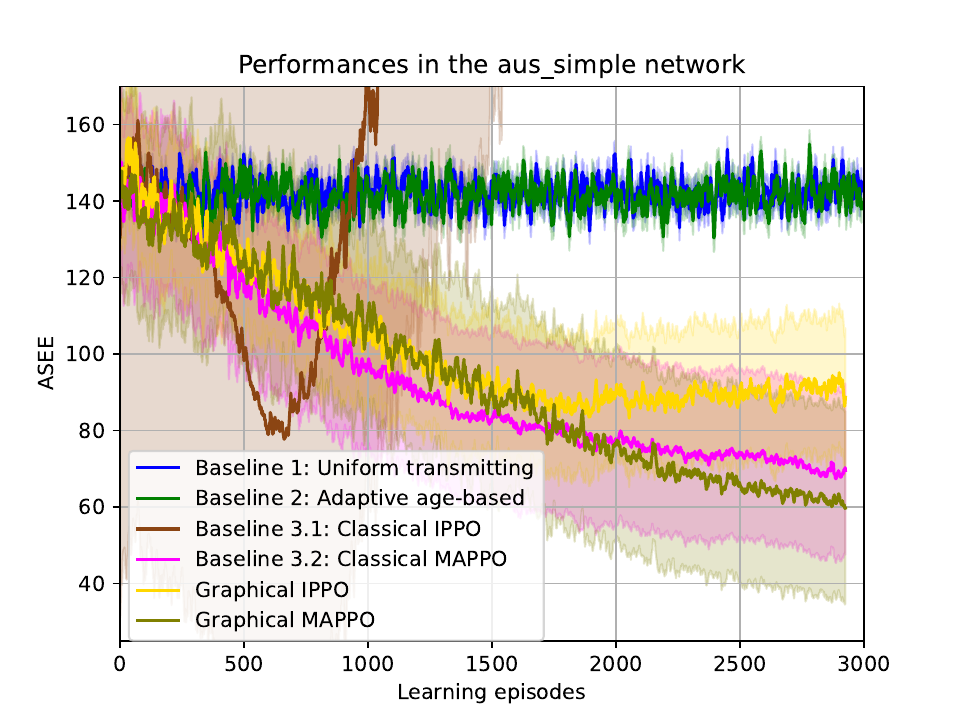}
    \caption{Real-data network}
\end{subfigure}
\caption{Performances of the proposed policies and baselines.}
\label{fig_performances}
\end{figure}

\textit{1) Performance on Synthetic and Real Networks}:
The ASEE results are shown in Fig.~\ref{fig_performances}(a) for Watts–Strogatz graphs and in \cite[Fig.~4(b)]{GNNerr} for stochastic block models. Key observations are: (i) Graphical policies consistently outperform their classical counterparts, demonstrating the benefit of incorporating graph structure.
(ii) Graphical MAPPO further outperforms graphical IPPO, confirming the advantage of CTDE over independent learning. (iii) Classical IPPO suffers from increasing estimation error due to the inherent non-stationarity of independent learning techniques, while graphical IPPO exhibits greater resilience to non-stationarity.

On the real network (Fig.~\ref{fig_performances}(c)), the trends are similar, though the advantage of graphical MAPPO is less pronounced: because the topology is fixed and the number of distinct permutations is limited. In such small networks, this reduces the benefit of graph-based learning. Nevertheless, after $3000$ episodes, graphical MAPPO still achieves a statistically significant ASEE improvement, with an average ASEE gap exceeding $10$ over the baselines.

\begin{figure}[htbp]
	\centering
	\begin{subfigure}{0.495\linewidth}
		\centering
		\includegraphics[width=1\linewidth]{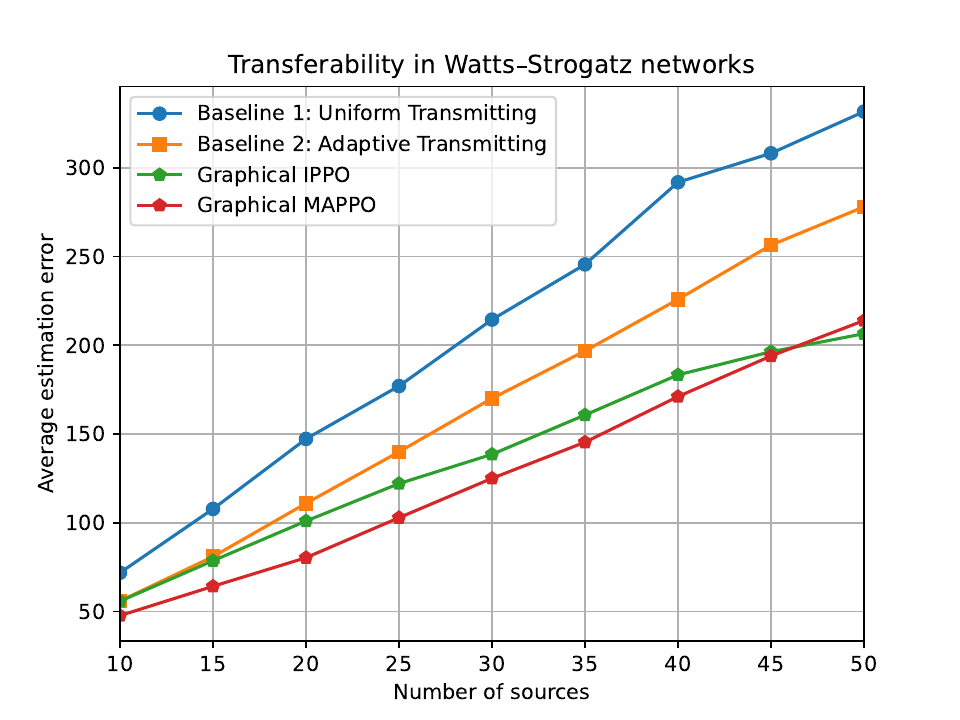}
		\caption{Transferability}
	\end{subfigure}
	\begin{subfigure}{0.495\linewidth}
		\centering
\includegraphics[width=1\linewidth]{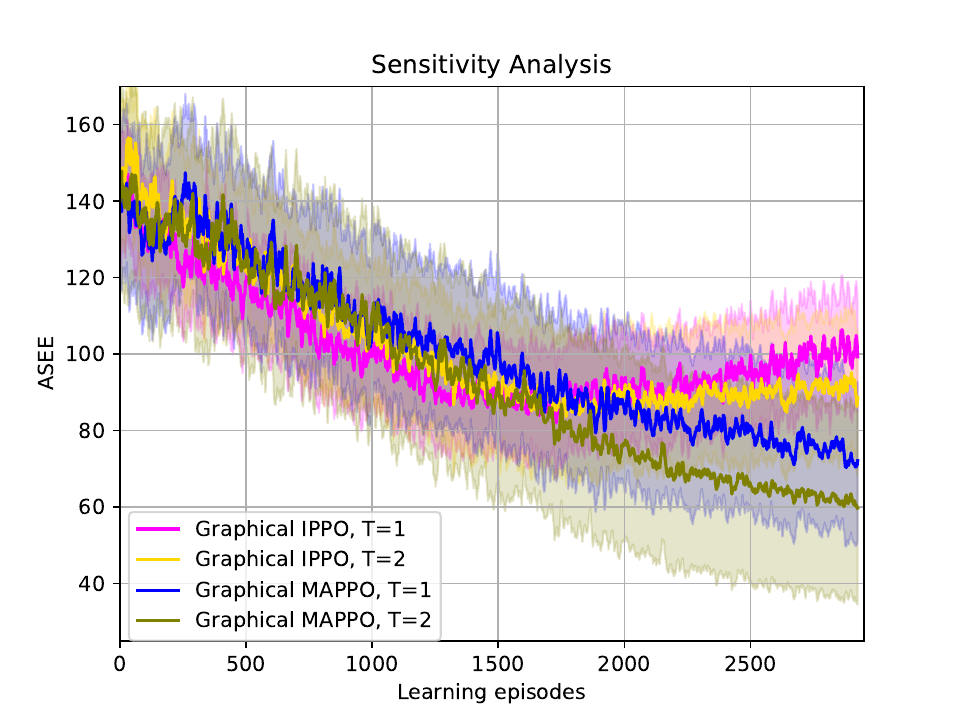}
\caption{Sensitivity Analysis}
	\end{subfigure}
	\caption{(a) Transferability in Watts–Strogatz networks: The policies are trained on $10$-node networks and tested on networks with $M\in[10, 50]$ nodes. (b) Performances of proposed policies in the real network under different $T$.}
	\label{fig_transferability_1} 
\end{figure}

\textit{2) Transferability}:
The transferability of our frameworks is shown in Fig.~\ref{fig_transferability_1}(a) and \cite[Fig.~5(b)]{GNNerr}. Models trained on $10$-node Watts–Strogatz networks and stochastic block models are applied to larger networks. As the number of nodes increases, the performance gap over the baselines widens, indicating that the advantages in small networks not only persist but are amplified in larger ones.

Interestingly, graphical IPPO policies eventually surpass graphical MAPPO policies (around $M\approx45$ in Fig.~\ref{fig_transferability_1}(a) and $M\approx40$ in \cite[Fig.~5(b)]{GNNerr}). This is because transferability is guaranteed only for GNNs built on graph filters \cite{Transferability}, whereas the MAPPO critic employs a different GNN architecture. We expect that adopting graph filter–based critics would restore MAPPO’s superiority across all network sizes.

\textit{3) Sensitivity Analysis}:
We compare ASEEs of policies with ($T=2$) and without ($T=1$) recurrence in Fig.~\ref{fig_transferability_1}(b).
Across both graphical IPPO and MAPPO, recurrent policies consistently achieve lower ASEEs, confirming the benefit of recurrence. For graphical IPPO, ASEEs without recurrence decrease initially but rise again, whereas with recurrence the same pattern appears at a much slower rate. This suggests that recurrence enhances resilience to non-stationarity.

\section{Conclusion}\label{sec: Conclusion}

In this paper, we study decentralized sampling and transmission policies to minimize time-average estimation error in dynamic multi-hop wireless networks. We propose a graphical MARL framework and formally prove its transferability. Extensive simulations on synthetic and real networks confirm both its effectiveness and transferability.

\clearpage
{\small
\bibliographystyle{IEEEtran}
\bibliography{Journal/references}
}

\end{document}